\documentclass[11pt,a4paper]{article}
\usepackage[T1]{fontenc}
\usepackage[utf8]{inputenc}
\usepackage{lmodern}
\usepackage[british]{babel}
\usepackage[margin=25mm]{geometry}
\usepackage{amsmath,amssymb,booktabs,graphicx,subcaption}
\usepackage[numbers,sort&compress]{natbib}
\usepackage[hidelinks]{hyperref}
\usepackage{microtype}

\graphicspath{{figures/}}
\title{Chess Position Recognition using Faster R-CNN}

\author{Paras Govind \quad Ognjen Arandjelovi\'c\\~\\
School of Computer Science\\
University of St Andrews\\
United Kingdom}

\date{}

\begin{document}
\maketitle

\begin{abstract}
Performing chess game position recognition solely from a single image of a three-dimensional board requires predicting the position and orientation of the board relative to the camera, the occupancy of squares and the piece type, which includes its colour. We propose an R-CNN-based framework with independent components for piece recognition and board geometry estimation, whose predictions are combined to reconstruct the position. For piece recognition, we adapt Faster R-CNN using a class-weighted objective and a deeper classification head. The detector operates directly on the input image, retaining alternative piece hypotheses that are subsequently refined using constraints on piece counts and square occupancy. For board detection, we introduce an octagonal arrangement of eight labelled boundary keypoints, predicted using the keypoint head of Mask R-CNN. These provide redundant correspondences for homography estimation and encode board orientation. The estimated homography maps representative points from the piece boxes to an $8\times8$ grid. On a synthetic dataset, the modifications to piece detection increase mean average precision from 61.59\% to 90.14\%. Of the predicted board keypoints, 97.11\% are within 1\% of the image diagonal of their labelled targets. Using
ground-truth piece boxes with the predicted homographies gives correct square
assignments for every test position. The complete framework recovers 76.61\%
of test positions exactly and 96.49\% with at most one incorrect square.
\end{abstract}

\section{Introduction}

Players rely on chess engines to analyse their games especially when it is
unclear what the best move is in a position or why a move they played was
inaccurate. Games played online are recorded automatically and can be directly
fed into a chess engine for evaluation. However, for an over-the-board game,
the position on a physical board has to be manually copied over to a digital
board by placing virtual pieces one by one. This is quite laborious and
error-prone considering the number of pieces on a board. Players can take a
photograph of a chessboard to save it for later but still have to manually
recreate the position on a virtual board. In this paper we propose a framework
capable of analysing images of chessboards and generating the corresponding
piece arrangement.

Treating chess recognition as an elementary task by having a single neural
network take an image as input and output one of 13 classes for the 64 squares,
the thirteenth class representing an empty square, is a direct approach to the
problem. This implies that the model has to learn everything required to
recognize chess positions, which includes localizing pieces, classifying them,
localizing the chessboard squares, determining the occupancy of those squares,
and finally mapping those occupancies to the corresponding location in the
final list of 64 squares. Environmental factors include camera quality,
lighting, camera angle, occlusion of the chessboard squares, occlusion of pieces
by other pieces, and board and piece colours. However, some subtasks of chess
recognition do not need a neural network to solve them and can be done using an
algorithm. For example, given labelled corners of the chessboard in an image,
mapping points on the board plane to their corresponding locations on a regular
grid can be done by computing a homography matrix. Hence, it makes sense to
break the problem down into smaller parts.

Chess recognition can be broken down into two independent tasks, namely piece
detection and board detection. Some approaches employ a pipeline of components,
with each component simplifying the problem for the next. This brings about a
potential issue with such a pipeline, which is errors being propagated forward.
For example, W\"olflein and Arandjelovi\'c~\cite{Wolflein2021} break the problem
down into four stages, where the board detection stages determine what areas of
the image piece detection is performed on. Failure to detect the board correctly
can result in the piece detection failing as well. We use a Faster R-CNN model
to both localize and classify pieces simultaneously, and a Mask R-CNN model
with keypoint detection to determine the chessboard's position and orientation.
Both models operate on the input image. The detected board is used to compute
a homography matrix that is then applied to representative points from the
predicted piece bounding boxes. This removes the dependence of piece detection
on a preceding board crop, although errors in either component can still affect
the final position.

We propose an R-CNN-based framework for chess position recognition with
contributions to both piece recognition and board geometry estimation.
For piece recognition, we develop a detector that operates directly on the
input image, using a class-weighted objective and a deeper classification
head to distinguish visually similar pieces despite their unequal
frequencies. Its predictions are subsequently refined using constraints
on piece counts and square occupancy. For board detection, we introduce
an octagonal arrangement of eight labelled boundary keypoints, designed
to provide redundant correspondences for homography estimation and to
address the sensitivity of outer-corner predictions to errors in the
detected board box. The keypoint labels also encode board orientation.
The two components operate independently on the input image, and their
predictions are combined through a homography to reconstruct the chess
position. We evaluate the modifications to piece detection, the accuracy
of geometric assignment, and the performance of the complete framework.

\section{Related work}

Performing chess recognition solely based on a single image of a
three-dimensional board requires predicting the position and orientation of
the board relative to the camera, the occupancy of squares and the piece type,
which includes its colour. In images of two-dimensional boards, the chessboard
is facing the camera. Board localization only involves identifying the corners
of the board and then dividing the square made by the corners into a grid of
equal squares, making use of the fact that the board is a flat square. When it
comes to detecting three-dimensional chessboards the same principles apply,
though detecting the board becomes more challenging. Perspective distortion
alters the shapes of the board and pieces in the image, introducing difficulties
in accurately identifying and localizing them. A system operating on a single
image also has to determine the piece types without relying on a previously
known position or a record of the moves played.

Czyzewski et al.~\cite{Czyzewski2020} used an iterative heat map generation
method to locate the board. The heat map depicts the probability that a
sub-area of the image contains the chessboard. The most confident sub-area is
cropped into, and the process is repeated until the board corners are
determined. Their system also uses information about chess positions to help
resolve piece recognition. Another approach was used by
Xie et al.~\cite{Xie2018Chess}, who used piece templates and distorted them
based on where the target piece is located on the board. The distorted
templates are then compared with the pieces to classify them. They compared
this oriented chamfer method with convolutional neural networks and showed
that the performance was very similar, with the limiting factor being a small
dataset of images.

W\"olflein and Arandjelovi\'c~\cite{Wolflein2021} implemented a system they
called Chesscog as a four-stage pipeline. A Canny edge detector is followed by
clustering to merge lines and a RANSAC-based algorithm to detect the board
lattice. The next step involves using a CNN to predict whether each square is
occupied or not, followed by another CNN to predict the piece class for the
occupied squares. This resulted in a model with a per-square error rate of
0.23\% on their synthetic test set. However, this pipeline structure is
susceptible to errors in earlier stages being propagated to the later stages.
Zhu et al.~\cite{Zhu2023} designed a view-invariant chess recognition model
that aimed to handle cases with low camera viewing angles as well as high ones.
This addresses a significant difficulty because lower viewing angles increase
occlusion among pieces and obscure more of the board.

Learned board geometry has also been combined with object detection.
Ranasinghe et al.~\cite{Ranasinghe2023} use a neural network to regress board
corners in ChessEye, then rectify and crop the board before detecting pieces
with YOLO. Masouris and van Gemert~\cite{Masouris2024}
introduced ChessReD, a dataset of real photographs taken with mobile devices,
and investigated recognition directly from images. Their results demonstrate
the difficulty of transferring conclusions from synthetic boards to photographs
with greater variation. More recently, Seytre~\cite{Seytre2026} has used
learned square queries with a vision transformer to predict board occupancy.
Our framework addresses piece recognition and board geometry through two
independent detection components. The piece detector combines a class-weighted
objective with a deeper classification head, and its output is refined using
constraints on the set of predicted pieces. The board detector uses eight
labelled keypoints in an octagonal arrangement to recover a homography and
board orientation. By detecting pieces in the full input image and assigning
them to squares afterwards, we can combine their visual classification scores
with the geometric and piece-count constraints when reconstructing a position.

\section{Piece recognition}

\begin{figure}[tbp]
\centering
\begin{subfigure}{.49\linewidth}
\includegraphics[width=\linewidth]{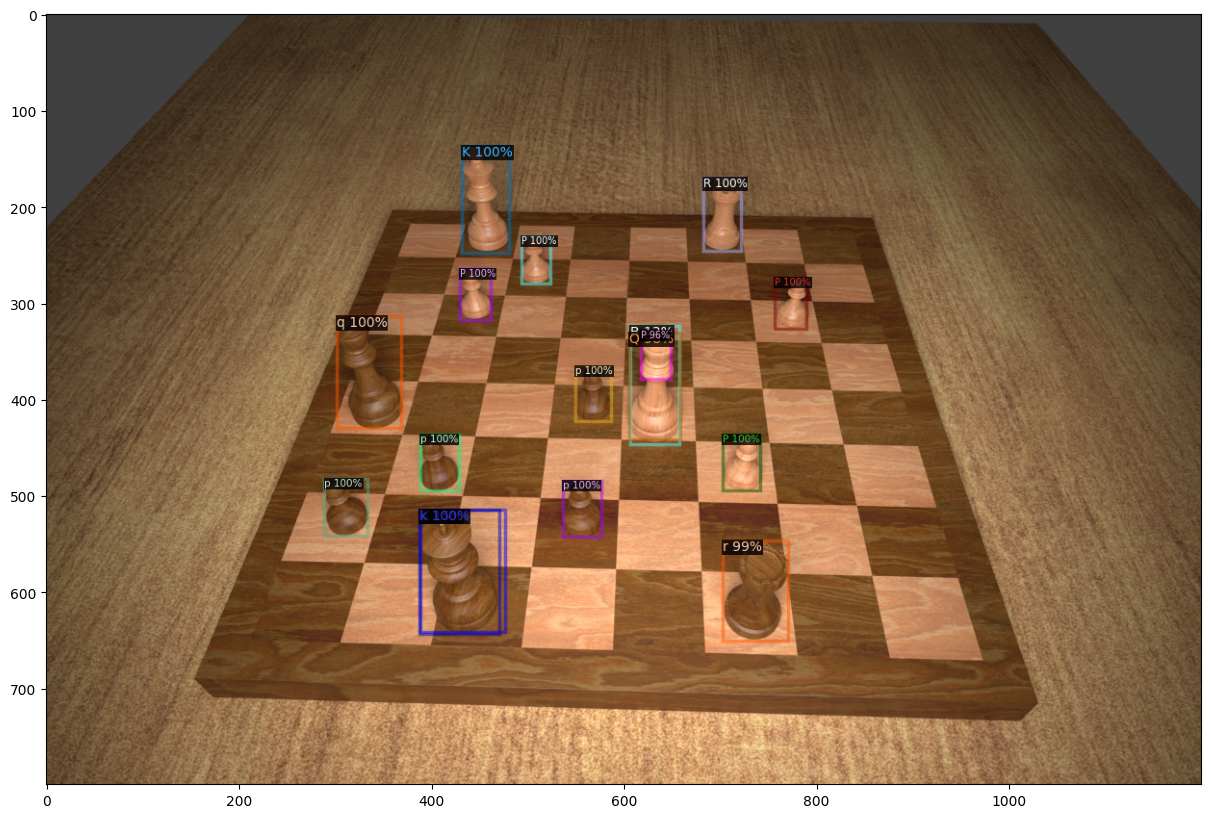}
\caption{Example with overlapping class hypotheses.}
\end{subfigure}\hfill
\begin{subfigure}{.49\linewidth}
\includegraphics[width=\linewidth]{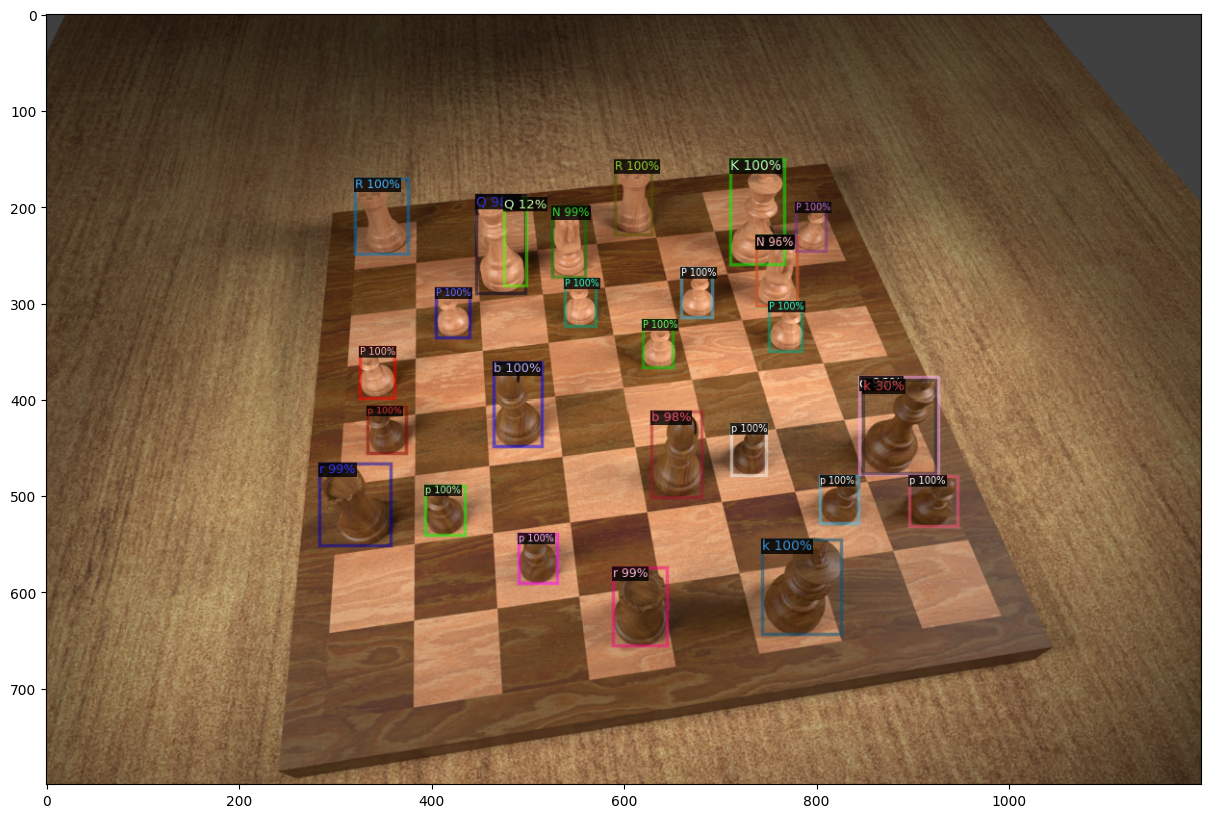}
\caption{Example under different illumination.}
\end{subfigure}
\caption{Unfiltered outputs of the final piece detector. Boxes are labelled
with predicted classes and confidence scores. Uppercase letters denote white
pieces and lowercase letters denote black pieces. Several hypotheses can be
retained for the same physical piece before the position is assembled.}
\label{fig:pieces}
\end{figure}

Piece detection consists of two parts, namely piece localization and piece
classification. Faster R-CNN~\cite{Ren2015} is a model that is able to do both
simultaneously. For a given input image, it outputs bounding boxes at locations
where it predicts pieces may exist, together with a class and confidence score
per bounding box. The model can predict several bounding boxes of different
classes for the same piece, as it only applies non-maximum suppression to
bounding boxes of the same class. Examples of the final detector's output are
shown in Figure~\ref{fig:pieces}. The small size of pieces and similarity
between pieces present challenges for classification, particularly for kings
and queens, which differ in relatively small details of shape.

Our piece-recognition procedure adapts this detector at both training and
prediction. During training, class weighting adjusts the contribution of
frequent and infrequent pieces to the classification loss, and a deeper box
head increases the capacity available for distinguishing their visual features.
At prediction, retaining several class hypotheses for a piece allows the
subsequent constraints to influence the final choice. Limits on the number of
pieces of each class are applied across the image. Once the board geometry
is available, square assignment provides a further means of resolving
competing predictions. The following subsections describe the detector
modifications and the piece-count constraints.

\subsection{Baseline model and class weighting}

Before modifying the model architecture significantly, it made sense to first
see how well the model performed piece detection with the least modifications.
We used Detectron2~\cite{Wu2019}, version 0.6, with a
ResNeXt-101-32x8d backbone~\cite{Xie2017} and a feature pyramid
network~\cite{Lin2017}. The region proposal network identifies candidate
regions, from which the box head predicts classes and refines bounding boxes.
The baseline box head has two fully connected layers and makes classification
predictions for each region separately.

The reason class frequencies are relevant to the training is because when
calculating cross-entropy loss, each class instance is weighted the same, hence
a higher frequency of instances leads to that class contributing more than
others towards the classification loss. Therefore, as the model is trying to
minimize the total loss, it can make larger reductions by improving predictions
for the high frequency classes. The dataset was generated from real games so
naturally there is an uneven distribution of pieces. In Table~\ref{tab:counts},
pawns make up over half of the instances. Queens have the lowest frequencies
because each side begins with only one queen and a queen can be captured during
the game. These frequencies motivate examining a weighted loss rather than
assuming that the unweighted objective gives comparable attention to all piece
types.

\begin{table}[tbp]
\centering
\caption{Piece counts in the augmented training data. Each image is
also reflected horizontally, so these counts are twice the unaugmented counts.}
\label{tab:counts}
\begin{tabular}{lrr}
\toprule
Piece & White & Black\\
\midrule
King & 8516 & 8516\\
Queen & 5494 & 5472\\
Rook & 12458 & 12474\\
Bishop & 9336 & 9306\\
Knight & 8390 & 8484\\
Pawn & 47182 & 47232\\
\midrule
Total & 91376 & 91484\\
\bottomrule
\end{tabular}
\end{table}

Our strategy involved the modification of the loss function by adjusting the
weights assigned to each class in the cross-entropy loss, considering their
respective frequencies. The rationale behind this approach is to assign a
relatively higher weight to classes with lower frequencies, enabling them to
exert a greater influence on the overall loss and compensating for their lower
occurrence. The weight assigned to each class is determined as the inverse of
its proportionate frequency in relation to the total frequency across all
foreground classes. That is, for each class its weight is the total frequency
divided by its frequency. For a foreground class $c$, with count $n_c$, the
class weight is
\begin{equation}
 w_c=\frac{N}{n_c}, \qquad N=\sum_{c=1}^{12} n_c,
 \label{eq:weights}
\end{equation}
and the corresponding classification loss contribution for a target of class
$c$ is $-w_c\log p_c$.

\subsection{Restricting piece counts}

The rules of chess apply several constraints to the locations and frequency of
pieces which can be used to better classify them. Given only the predicted
bounding boxes together with their class and confidence, there is a limit to
the rules that can be applied to them as most rules rely on the position of the
pieces relative to the board. A restriction that can be implemented at this
stage is a limit on the frequency of each piece class. For example, there can
only be one king and at most eight pawns on the board per colour. Barring pawn
promotion, there can be at most one queen, two rooks, two knights and two
bishops per colour. We use these limits under the assumption that there have
been no promotions. This assumption matches the positions used in the
evaluation. For positions with promotions, the limits would need to
accommodate additional queens, rooks, bishops or knights~\cite{FIDE2023}.

The process involves only keeping the most confident set of predictions for
each class that satisfy the constraints. For example, in the case of the black
queen, since we have assumed there can only be one instance of it, only the
bounding box with the highest black queen confidence is kept and all others
are discarded. Note that the application of this rule relies on the model
giving the true positive bounding box prediction a higher confidence score
than a false positive, otherwise the post-processing step can eliminate true
positive boxes. For this class frequency rule, the prediction boxes of each
class are independent of other classes, hence the order of classes to which
the rule is applied does not change the result. The resulting predictions
satisfy the chosen upper limits on piece counts.

\subsection{Deepening the classification head}

The baseline box head is made of two fully connected layers responsible for
extracting high level features. By increasing the depth of the network we aim
to increase the capacity of the model to capture more complex features to
distinguish between chess pieces more effectively. A deeper network may extract
finer details of these pieces, which may result in improved classification
performance. With a deeper model, we have to be careful about overfitting
occurring, as this extra capacity can instead be used to memorize the training
dataset rather than learning general features.

To address this, we decided to start with a deep network with several
additional hidden layers and evaluate its performance on the validation set.
A layer would then be pruned, and the new model would be re-evaluated. If the
degradation in performance was minimal, another layer would be pruned, and
the process would be repeated until a significant drop in performance was
observed. The models were evaluated without applying the class frequency rule
as the aim was to find a model that performed well independently. The custom
loss function was still used. The deepest model had five fully connected
layers in the box head. The pruning was done until it was observed that with
three layers the model's performance dropped substantially. Hence, the model
with four layers was chosen as the final one.

\section{Board geometry and position reconstruction}

Simply knowing all the pieces that are present on a chessboard is not enough
to define a position, as to do that we are required to place each piece on a
square on the board. From the piece bounding boxes it may be possible to
determine where pieces are relative to each other, but we still need to
determine where the pieces are relative to the board. Hence board detection
is a necessary step in the chess recognition pipeline. The bounding box around
a chessboard in an image is not enough to determine its exact position in the
image. Previous representations include identifying the four corners of the
board, its outer edges, or a grid of lattice points representing the boundaries
of its squares. All these representations enable the localization of the pieces
with respect to the board, which ultimately determines the final chess position.
In the proposed framework, the board component predicts labelled keypoints
from which we recover the transformation needed to assign the piece detections
to squares.

\subsection{Homography and labelled keypoints}

\begin{figure}[tbp]
\centering
\includegraphics[width=.92\linewidth]{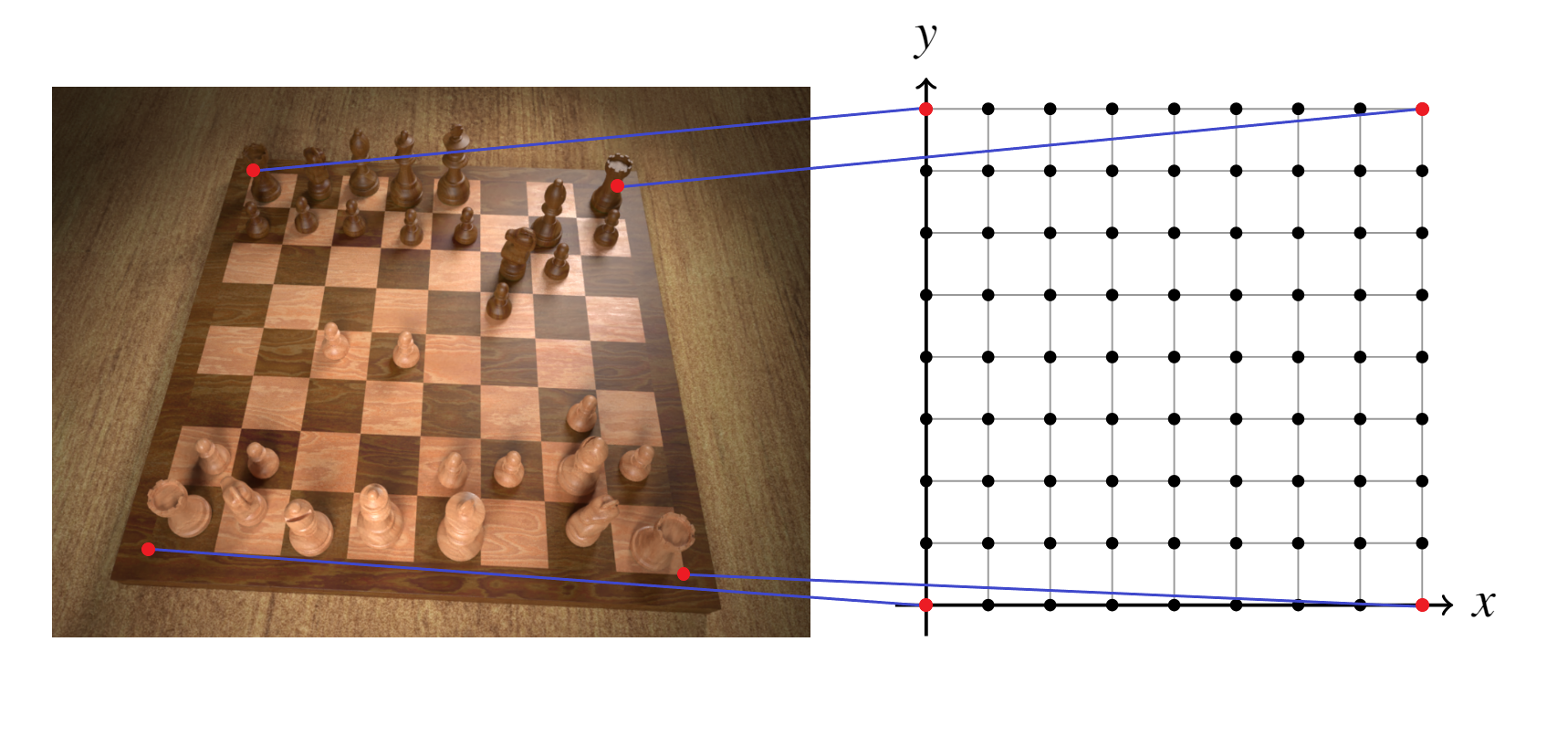}
\caption{Projection from an image of a chessboard to a regular grid. Labelled
correspondences determine both the board geometry and its orientation in the
grid. The transformation applies to the board plane.}
\label{fig:homography}
\end{figure}

Suppose a chessboard in an image has been detected by its corners. Given the
pixel coordinates of the corners and the pieces, the task is now to determine
where each piece is relative to the board. The end goal is to assign one of
64 squares to each piece on the board, which would be our final position.
If the board was aligned with the camera then the solution would involve
dividing the square defined by the detected corners into an $8\times8$ grid
and checking which grid square each piece belongs to. However, in our problem
the board may not be aligned with the camera, hence the pixel coordinates
cannot be used directly without accounting for perspective distortion. The
solution, visualized in Figure~\ref{fig:homography}, is to project points on
the chessboard plane to a flat regular grid using a homography matrix.

The transformation is represented as a $3\times3$ matrix $H$, defined up to a
nonzero scale. To derive such a matrix, at least four corresponding points
are needed, with no three of a minimal four-point set collinear. In homogeneous
coordinates, an image point $p=(x,y,1)^\mathsf{T}$ and its grid location
$q=(u,v,1)^\mathsf{T}$ satisfy $q\sim Hp$. Writing the rows of $H$ as
$h_1^\mathsf{T}$, $h_2^\mathsf{T}$ and $h_3^\mathsf{T}$, the inhomogeneous
projection is
\begin{equation}
 \pi_H(p)=\left(
 \frac{h_1^\mathsf{T}p}{h_3^\mathsf{T}p},
 \frac{h_2^\mathsf{T}p}{h_3^\mathsf{T}p}\right),
 \label{eq:homography}
\end{equation}
provided the denominator is nonzero. In the regular grid, the side closer to
the horizontal axis is fixed to be White's side. Given the pixel coordinates
of the corners, mapping the image chessboard to the regular grid requires
knowing which corners are supposed to be projected to which corners in the
grid. This is how orientation is expressed here. If the camera faces the
chessboard from Black's direction, the near corners are projected to the far
corners of the grid.

The solution we devised involves using Mask R-CNN for keypoint
detection~\cite{He2017}, enabling the identification of specific points on
the board. Keypoints denote points of interest within the object depicted in
an image. In our case, these represent specific locations on the chessboard.
Each keypoint is associated with a distinct label, and by assigning these
labels to particular points on the regular chess grid, we determine the
orientation of the chessboard depicted in the image. The model therefore
predicts both image locations and their correspondence to the canonical board.
The masks are not needed in this case as the keypoints are the output we
require. The remaining design consideration is the selection of these
keypoints.

\subsection{From a cross to an octagon}

With accuracy in mind, it is sensible to choose locations such as the corners
of squares or the edges between squares, rather than, for example, the centre
of a square. The rationale behind this choice lies in the fact that the
centres of squares lack distinctive features that can be directly identified.
To predict the centre accurately, the model would need to analyse a wider
area of the image and first detect the boundaries of the squares. Hence by
setting the boundaries themselves as keypoints we aim to make the problem
simpler for the model. Furthermore, corners are more identifiable than
arbitrary points along edges as they are an intersection of two boundaries.
Therefore, a decision was made to use square corners as keypoints.

In the entire chessboard there are a total of 81 such corners. If the
chessboard was empty then there would be little difference in the difficulty
of predicting any of them. However, with pieces on the board, some corners
are more likely to be occluded by pieces standing between the camera and the
corner. A corner hidden from view will carry greater uncertainty regarding
its location in the image than a visible corner. Therefore, the rationale is
to select corners that have a low likelihood of being occluded by pieces.
Boundary locations are natural candidates, although their visibility still
depends on the position and viewpoint.

\begin{figure}[tbp]
\centering
\begin{subfigure}{.40\linewidth}
\includegraphics[width=\linewidth]{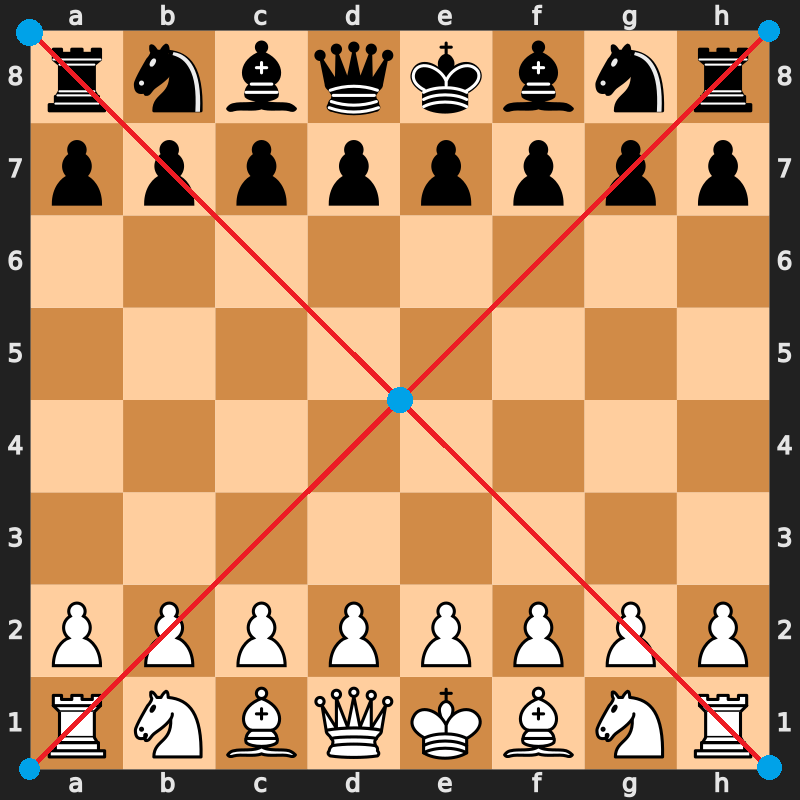}
\caption{Four outer corners and the centre.}
\end{subfigure}\hspace{.05\linewidth}
\begin{subfigure}{.40\linewidth}
\includegraphics[width=\linewidth]{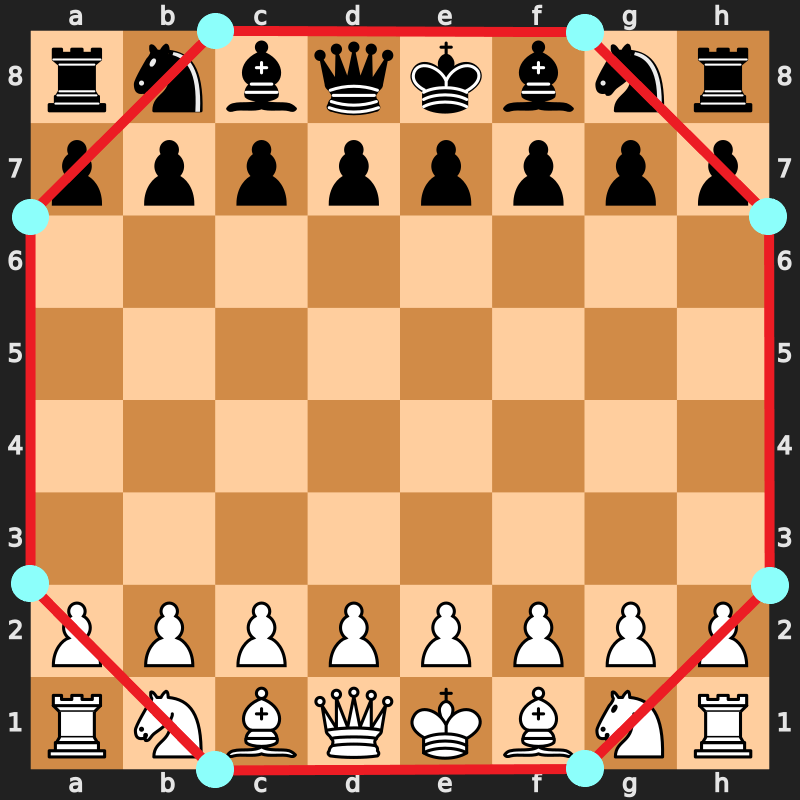}
\caption{Eight boundary keypoints.}
\end{subfigure}
\caption{The two keypoint layouts investigated. Blue points mark labelled
locations. In the cross layout, the centre is collinear with each pair of
opposite corners. In the octagonal layout, no three selected points are
collinear. The connecting lines illustrate the layouts and are not additional
predicted features.}
\label{fig:layouts}
\end{figure}

Given that the computation of the homography matrix requires a minimum of
four points, detecting all four corners of the chessboard meets the numerical
requirement. However, it is beneficial to have more points as they can be used
for validation purposes. The initial choice was to pick the centre of the
board as a fifth point as it sits between the four corners. If we connect
each corner to the centre with a line, we get a cross shape on the regular
grid, as shown in Figure~\ref{fig:layouts}. The validation method entails
computing the homography matrix from the four corners and then calculating
its inverse. This inverse matrix is used to project the centre of the regular
grid onto the image chessboard and compare it with the predicted location of
the fifth point. Agreement within a threshold indicates consistency with the
four corners.

\begin{figure}[tbp]
\centering
\includegraphics[width=.73\linewidth]{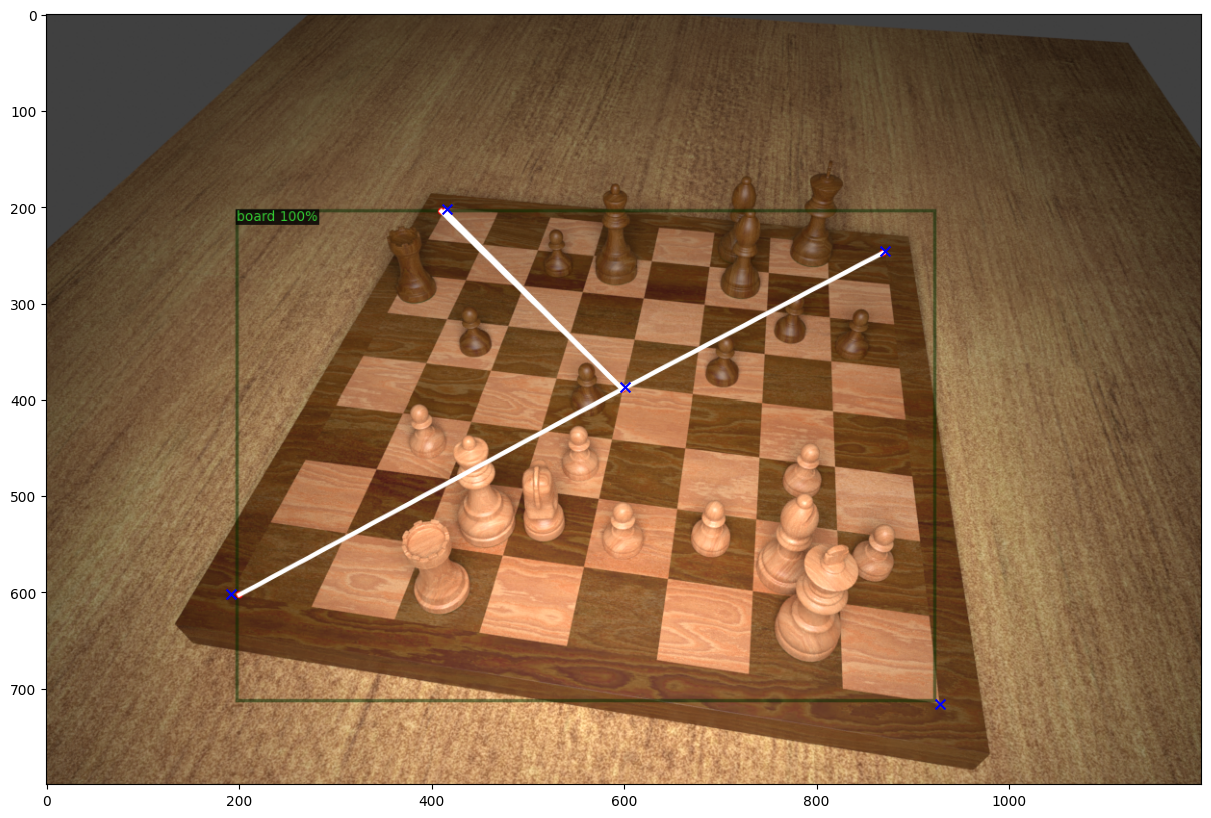}
\caption{Failure of the cross layout when the predicted board box excludes
an outer corner. The keypoint head cannot place a prediction outside the box.
The blue crosses mark true locations, and the connected predictions show the
resulting anomalous keypoint.}
\label{fig:cornerfailure}
\end{figure}

A keypoint detection model was trained to detect the board as the only class
together with five keypoints. This model is set up to predict a location for
all keypoints within the detected bounding box of the board. On inspecting
its predictions, we observed instances where the predicted bounding box would
crop into the actual chessboard, resulting in a corner of the board being
outside the box. Figure~\ref{fig:cornerfailure} shows one such instance.
This results in the model being unable to place the keypoint at the board
corner as it is restricted to the space inside the bounding box. The predicted
keypoint can end up being placed near another corner, which results in a
corner having two keypoints in its area.

The problem with choosing the centre as the fifth point is that it is
collinear with both pairs of opposite corners. Any selection containing the
centre and three outer corners therefore contains three collinear points and
cannot determine a general homography. The centre can check a transformation
computed from all four outer corners, but cannot replace a missing corner in
a minimal estimate. Nevertheless, training the model to detect a cross
identified a potential problem with the bounding box detection. The true
bounding box for the chessboard is a tight fit such that each side of the box
touches at least one corner of the board. Hence if a predicted box edge moves
inwards, a corner may be excluded. The outer corners are therefore vulnerable
to small errors in the bounding box prediction.

These observations led to the final keypoint selection strategy. The next
candidates are the square corners on the edges of the board. Another change
to the cross is to add more points so that the transformation can be estimated
after rejecting inconsistent predictions. The proposed keypoint selection is
a set of vertices in the shape of an octagon, as shown in
Figure~\ref{fig:layouts}. On a board spanning $[0,8]\times[0,8]$, the
selected locations are
\begin{equation}
\mathcal{K}=\{(2,0),(6,0),(8,2),(8,6),
             (6,8),(2,8),(0,6),(0,2)\}.
\label{eq:octagon}
\end{equation}
No three of these points are collinear, so every four-point subset is
geometrically admissible for a homography estimate. Under an oblique view,
these locations can also lie inside the axis-aligned board box with some
margin to its edges. This reduces the particular vulnerability of outer
corners. The available margin depends on the orientation of the board
relative to the image axes.

RANSAC~\cite{Fischler1981} is used to determine outlier keypoints and produce
a homography from the remaining points. It works by fitting models to subsets
of the data and selecting the model with the largest consensus set. For a
planar homography, each minimal sample contains four correspondences. The
estimated transformation is tested on the other points, and predictions
consistent within a threshold are treated as inliers. The final transformation
is then refined using the inliers. The rationale is that inliers produce
models that agree with other inliers, whereas outliers generally do not
produce a transformation consistent with many other points.

\subsection{Combining piece and board predictions}

\begin{figure}[tbp]
\centering
\begin{subfigure}{.48\linewidth}
\centering
\includegraphics[height=5cm]{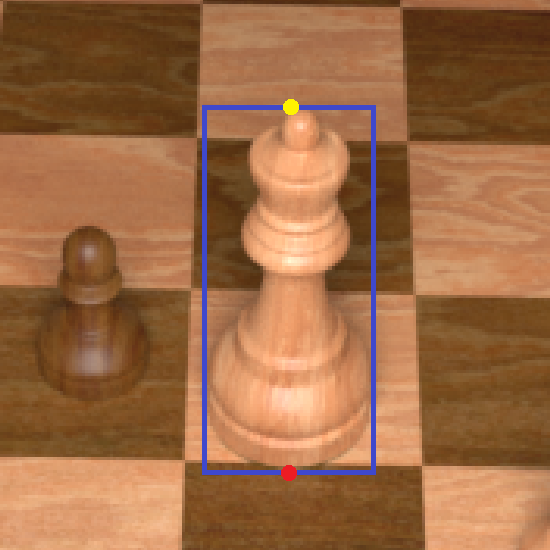}
\caption{A queen whose base crosses a square boundary.}
\end{subfigure}\hfill
\begin{subfigure}{.48\linewidth}
\centering
\includegraphics[height=5cm]{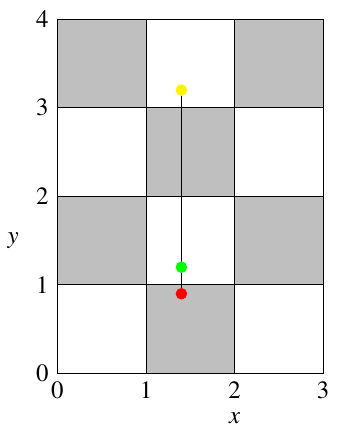}
\caption{The corresponding projected points.}
\end{subfigure}
\caption{Selection of a representative point for square assignment. The red
point is the bottom midpoint of the box and the yellow point is its top
midpoint. After projection, the green point is displaced by 0.2 square units
from the red point towards the yellow point.}
\label{fig:anchor}
\end{figure}

The two recognition components are combined by applying the estimated
homography to representative points from the detected piece boxes. The
detector provides each piece hypothesis and its confidence, and the board
component provides the transformation to labelled squares. Since each
predicted piece is represented as a bounding box, the remaining task is to
select a specific point to represent it. The initial selection strategy was
to use the bottom middle of the bounding box as the point for the piece.
The problem with this approach is that there may be instances where part of
a piece is standing over an edge, such as in Figure~\ref{fig:anchor}. The
bottom middle of the bounding box can be on the adjacent square, which
implies that the piece will get projected to the wrong square even though
most of its base is on one square. The piece base refers to the circular
portion of each piece that is in contact with the board.

Hence the final strategy is to try to approximate the centre of the piece
base. To do this we assume that from the camera's point of view, the centre
of the piece base is on the line connecting the bottom middle and top middle
of the bounding box. We then make another assumption by treating the piece
bases as circles with a small radius relative to the square size. The process
involves projecting the bottom middle and top middle points of each bounding
box and then picking a point along that line that is 0.2 grid units away from
the projected bottom middle. Writing these two projected points as $b$ and
$t$, respectively, the representative point is
\begin{equation}
 r=b+0.2\,\frac{t-b}{\lVert t-b\rVert_2},
 \label{eq:anchor}
\end{equation}
when $t\ne b$. This fixed displacement approximates the centre of the piece
base for square assignment. The board homography acts on the board plane,
and the projected top midpoint is used to determine the direction of the
displacement. We evaluate this heuristic through the resulting square
assignments.

Each representative point is assigned a square based on its coordinates in
the regular grid. Since there can only be one piece per square, the bounding
box with the highest confidence is selected as the final piece for each
square. Additional rules that rely on the position of pieces on the board
are applied, such as pawns only existing between the second and seventh ranks.
Together with the class frequency limits, these operations discard incompatible
predictions. The resulting arrangement can be written as the piece-placement
field of Forsyth--Edwards Notation~\cite{Edwards1994}. A single image does not
generally reveal whose turn it is, castling rights, or the preceding move
needed to determine an en passant target, so those fields cannot in general
be recovered from the visible arrangement alone.

\section{Experimental evaluation}

We evaluate the two components separately and then assess the complete
recognition framework. The piece-recognition experiments compare the baseline
detector with class weighting, piece-count filtering and the deeper
classification head. For the board component, we measure keypoint and
reprojection errors, followed by square-assignment accuracy using true piece
boxes. Finally, we combine the predicted boxes and board geometry to evaluate
the reconstructed positions. This separates the effects of the detector
modifications from the geometric assignment of pieces to squares.

\subsection{Dataset and training}

We used the dataset W\"olflein and Arandjelovi\'c~\cite{Wolflein2021} put
forth to train our models. It is a synthetic dataset rendered in Blender.
A benefit of this dataset is that the generated chessboards are positions
from real games played by expert chess players. This means that the positions
are not random arrangements of pieces but follow patterns arising in play.
Moreover, since the images were rendered from known chess positions, the
identities of the pieces and their locations on the board were known exactly.
The corresponding ground-truth annotations were therefore generated
automatically without uncertainty in the labels. The entire dataset
has 4,888 images. The images have variation in spotlight intensities and
angles, allowing the models to be evaluated under these changes in lighting.
Horizontal reflection was used to augment the training data.

The dataset uses the same colour scheme and piece appearance throughout,
with the camera angle ranging between $45^\circ$ and $60^\circ$ with
respect to the board. Images are taken from either White's side or Black's
side of the board. These conditions define the range of viewpoints and
appearances considered in the experiments. Within this setting, variation
in the positions and illumination allows us to evaluate the framework under
different configurations of occlusion and lighting.

We divided the dataset into training, validation and test sets. The test set
contains 342 images and 7,387 pieces. The keypoint layout was refined through
visual inspection of predictions, including predictions on test images.
The baseline piece model was trained for 5,000 iterations with a batch size
of four and a base learning rate of 0.001. Validation was performed every
250 iterations. To allow a comparison with the baseline model, the weighted
two-layer model was trained for the same number of iterations. The initial
five-keypoint board model was trained for 3,000 iterations.

\subsection{Piece detection results}

Generating a classification confusion matrix involves associating each
predicted bounding box with its corresponding ground-truth piece. This was
done by categorizing prediction boxes into buckets for each ground-truth
piece, determined by the intersection over union of the predicted and true
boxes. Prediction boxes with an IoU of at least 70\% were allocated to the
respective buckets. Given that the model can predict multiple bounding boxes
for each piece, the bounding box with the highest confidence was selected
from each bucket to serve as the final prediction for that piece. Hence these
confusion matrices evaluate classification conditional on a piece having been
localized. Missed pieces and unmatched predictions must be considered
separately. We also evaluate detection and classification together using
per-class average precision and its mean across the 12 piece classes.

Out of the 7,387 pieces in the test dataset, the baseline model localized
7,335, and there were 399 prediction bounding boxes that did not overlap
sufficiently with any piece. Among the localized pieces, 6,144 were correctly
classified, resulting in a classification accuracy of 83.76\%. The model
performed well for pawns, rooks and kings of both colours but poorly for
knights, bishops and queens. It frequently predicted queens to be kings,
which is consistent with the similarity in their appearance. This also
reduced the AP for the king classes despite their high recall in the
conditional confusion matrix. The mAP of the baseline model was 61.59\%.

\begin{figure}[tbp]
\centering
\begin{subfigure}{.32\linewidth}
\includegraphics[width=\linewidth]{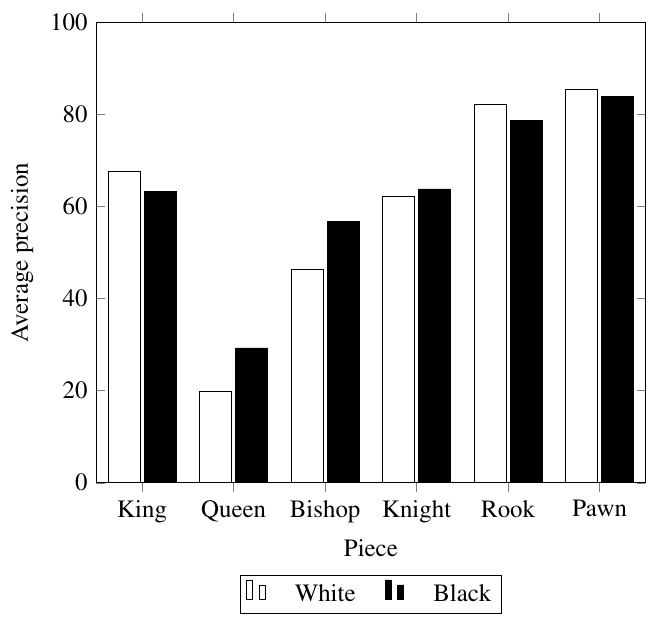}
\caption{Baseline.}
\end{subfigure}\hfill
\begin{subfigure}{.32\linewidth}
\includegraphics[width=\linewidth]{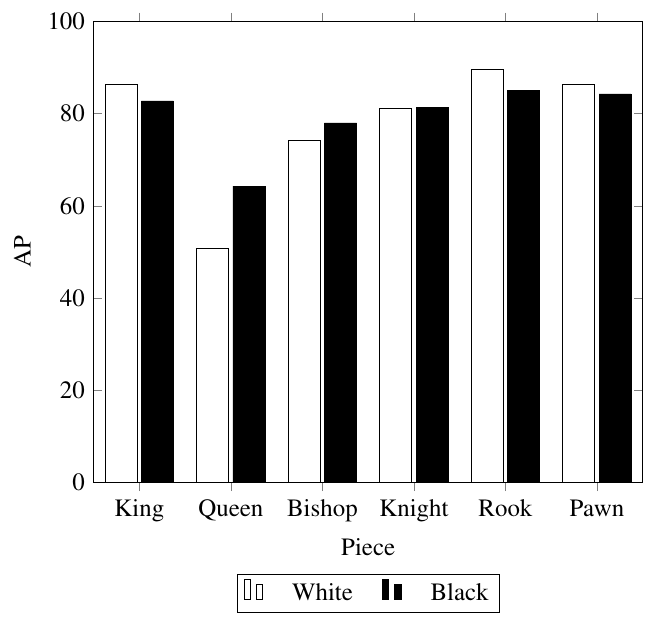}
\caption{Class-weighted loss.}
\end{subfigure}\hfill
\begin{subfigure}{.32\linewidth}
\includegraphics[width=\linewidth]{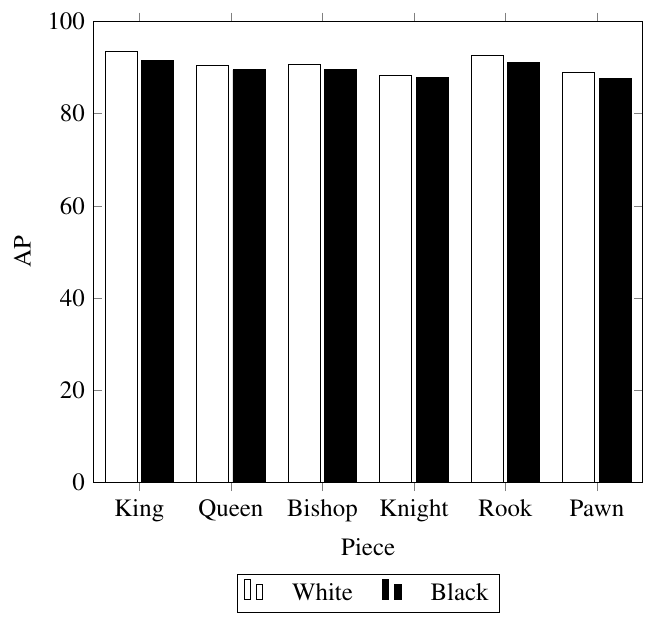}
\caption{Weighted, four-layer head.}
\end{subfigure}
\caption{Per-class average precision for the three piece detectors. The mAP
values are 61.59\%, 78.64\% and 90.14\%, respectively.}
\label{fig:ap}
\end{figure}

The custom loss function model shows an improvement in the precision of all
classes, with the largest changes for queens, bishops and knights
(Figure~\ref{fig:ap}). It attained a mAP of 78.64\%, an increase of
17.05 percentage points from the baseline model. The model localized
7,352 pieces correctly but had 1,285 unmatched predictions. Of the localized
pieces, 6,715 were classified correctly, giving a classification accuracy of
91.34\%. Therefore, the improvement in conditional classification came with
an increase in false detections. Table~\ref{tab:detector} reports these
quantities separately so as to preserve this distinction.

\begin{table}[tbp]
\centering
\small
\caption{Piece detection diagnostics on 7,387 ground-truth pieces. Correct
classification is measured only among matched pieces. The comparison shows
the effects of class weighting and piece-count filtering on the matched
pieces and unmatched predictions.}
\label{tab:detector}
\begin{tabular}{lrrrr}
\toprule
Configuration & Matched & Unmatched & Correctly & Conditional\\
 & pieces & predictions & classified & accuracy\\
\midrule
Baseline & 7335 & 399 & 6144 & 83.76\%\\
Weighted loss & 7352 & 1285 & 6715 & 91.34\%\\
Weighted loss + count limits & 7212 & 540 & 7002 & 97.09\%\\
\bottomrule
\end{tabular}
\end{table}

\begin{figure}[p]
\centering
\begin{subfigure}{.49\linewidth}
\includegraphics[width=\linewidth]{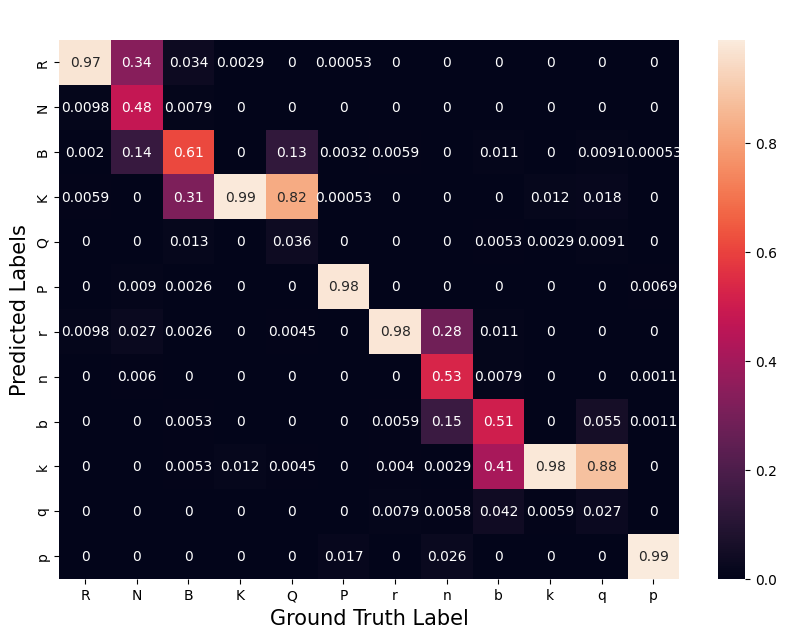}
\caption{Baseline.}
\end{subfigure}\hfill
\begin{subfigure}{.49\linewidth}
\includegraphics[width=\linewidth]{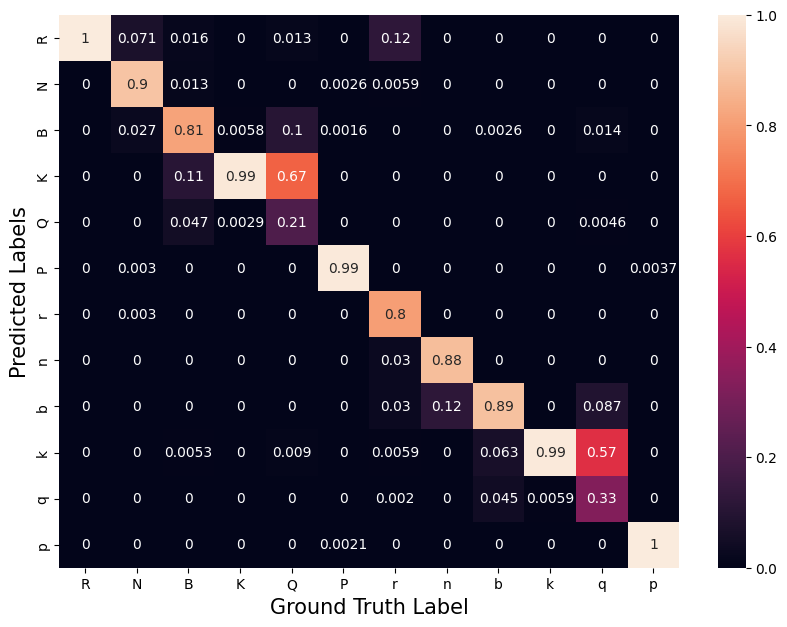}
\caption{Class-weighted loss.}
\end{subfigure}

\medskip

\begin{subfigure}{.49\linewidth}
\includegraphics[width=\linewidth]{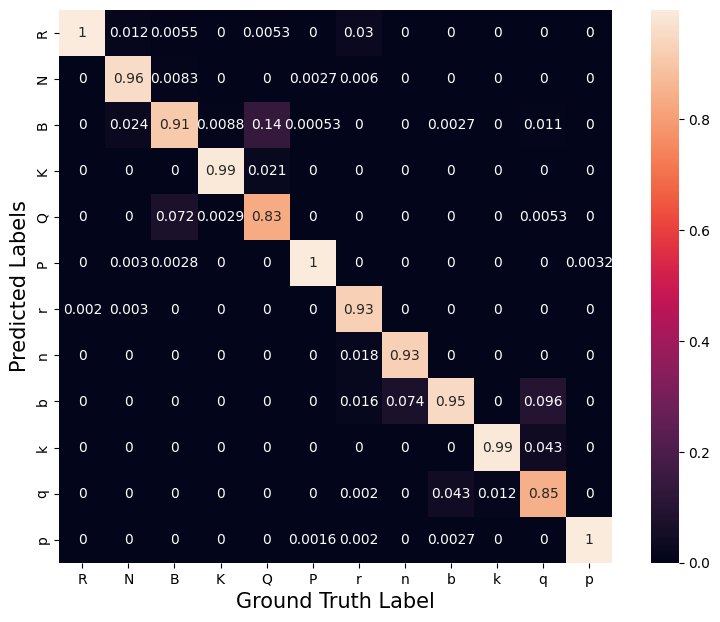}
\caption{Weighted loss with count limits.}
\end{subfigure}\hfill
\begin{subfigure}{.49\linewidth}
\includegraphics[width=\linewidth]{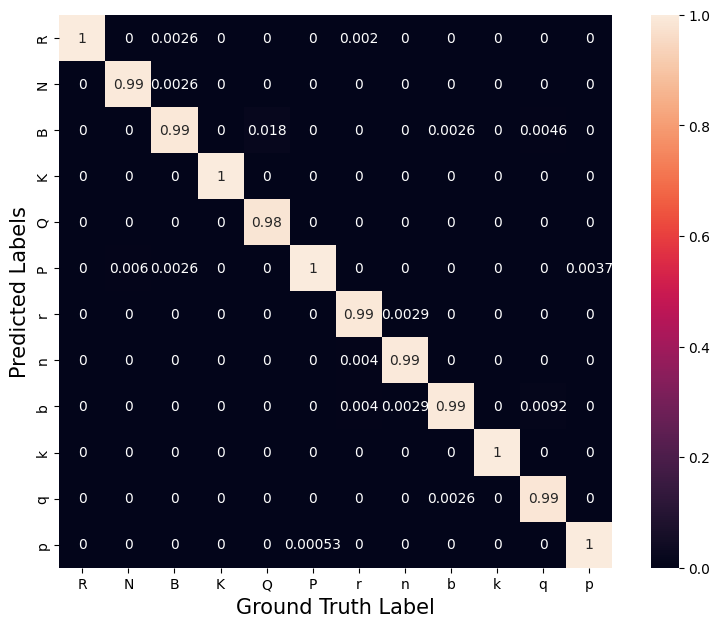}
\caption{Four-layer head with count limits.}
\end{subfigure}
\caption{Classification confusion matrices for matched pieces.
Columns are true classes and rows are predicted classes, with each column
normalized by its number of matched instances. Uppercase letters denote
white pieces and lowercase letters denote black pieces. These matrices
exclude missed pieces and unmatched detections.}
\label{fig:confusions}
\end{figure}

Applying the class frequency rule improved the classification of queens,
bishops and knights (Figure~\ref{fig:confusions}). The post-processing step
reduced the matched piece total to 7,212 and increased the missed piece total
to 175. However, the unmatched prediction total fell to 540. The model now
correctly classified 7,002 of the matched pieces, resulting in a conditional
classification accuracy of 97.09\%.

The four-layer model improved classification further. With the class frequency
rule applied, its overall conditional classification accuracy was 99.02\%.
The AP of each class improved, with a more balanced AP across the classes
and a minimum of 87.61\%. The mAP of the deep model with the custom loss
function was 90.14\%. These results motivate the combination of class weighting
and the deeper classification head in the piece-recognition component of
the framework.

\subsection{Board localization and square assignment}

\begin{figure}[tbp]
\centering
\begin{subfigure}{.49\linewidth}
\includegraphics[width=\linewidth]{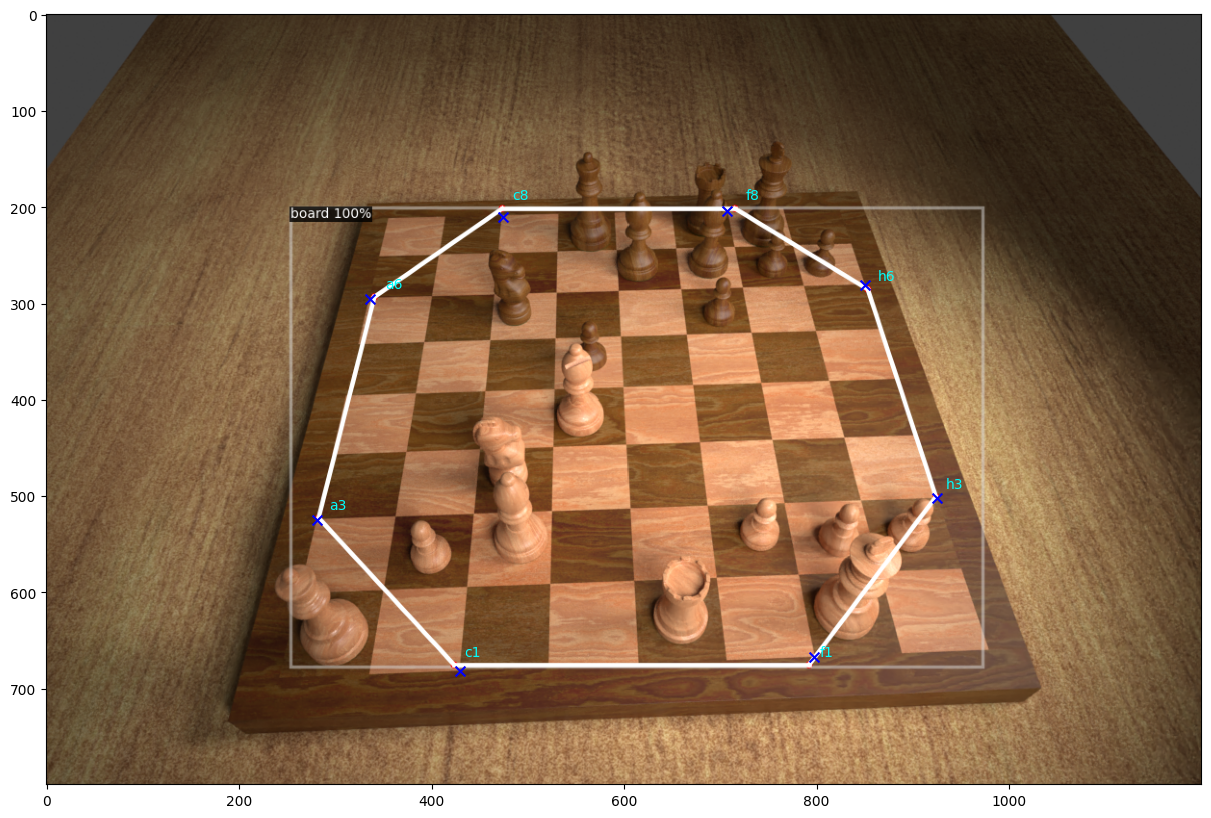}
\caption{First example.}
\end{subfigure}\hfill
\begin{subfigure}{.49\linewidth}
\includegraphics[width=\linewidth]{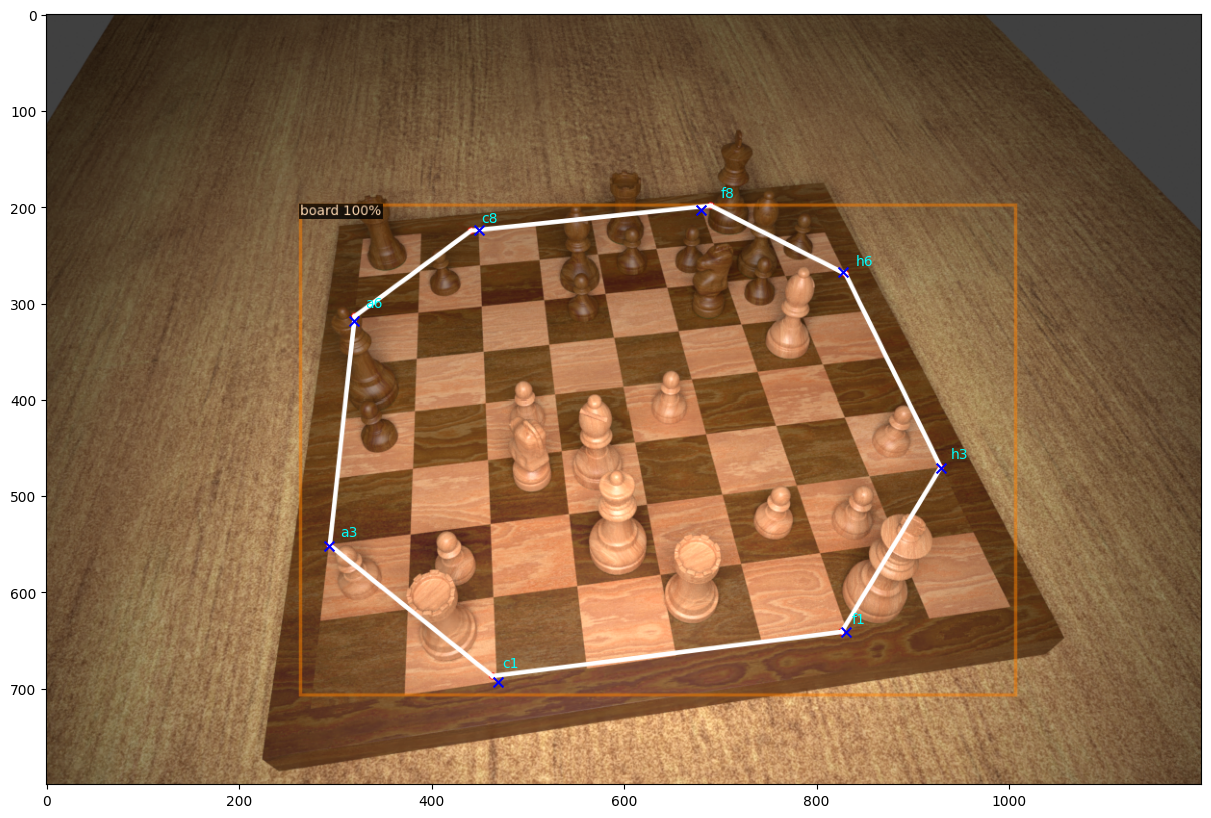}
\caption{Second example.}
\end{subfigure}
\caption{Outputs of the final board detector. The eight labelled predictions
are connected to show their order around the board. The predicted enclosing
box is also shown. Some board features are occluded by pieces.}
\label{fig:boardexamples}
\end{figure}

The board model was evaluated by measuring the distance between the predicted
and true keypoint locations. Since each keypoint is labelled, it is compared
with the true location of its corresponding keypoint. This implies that even
if the model detects all the points on the chessboard correctly, assigning
their labels incorrectly can produce a large error. The error between the
predicted and true point is calculated as the Euclidean distance between the
pixel coordinates of the two points divided by the image diagonal. The
normalization factor makes the measurement relative to image size. On the
test set, 97.11\% of predicted keypoints had an error of less than 1\%.
To put this to scale, 1\% corresponds to approximately 14.4 pixels in a
$1200\times800$ image. Sample predictions are shown in
Figure~\ref{fig:boardexamples}.

To evaluate the transformation produced from these keypoints, the 81 true
square corners in each test image were projected to the regular grid using
the estimated homography. In the regular grid, the Euclidean distances
between the projected points and the true corners were measured. Those
distances were squared and then averaged, giving a mean squared
reprojection error of 0.06, with each square having unit side length. This
quantity is measured in squared grid units, and the corresponding root mean
squared error is approximately 0.245 square units.

A further evaluation used the true piece bounding boxes and classes in
conjunction with the predicted homography matrix to reconstruct chess
positions. The piece-coordinate rule in Equation~\ref{eq:anchor} was applied
to each box. The resulting square assignments were correct for every test
and validation position. The training success rate was 99.74\%, showing that
the predicted geometry was sufficient for assignment with the true boxes
in those test images.

\subsection{Complete positions and comparison with Chesscog}

\begin{figure}[tbp]
\centering
\begin{subfigure}{.58\linewidth}
\includegraphics[width=\linewidth]{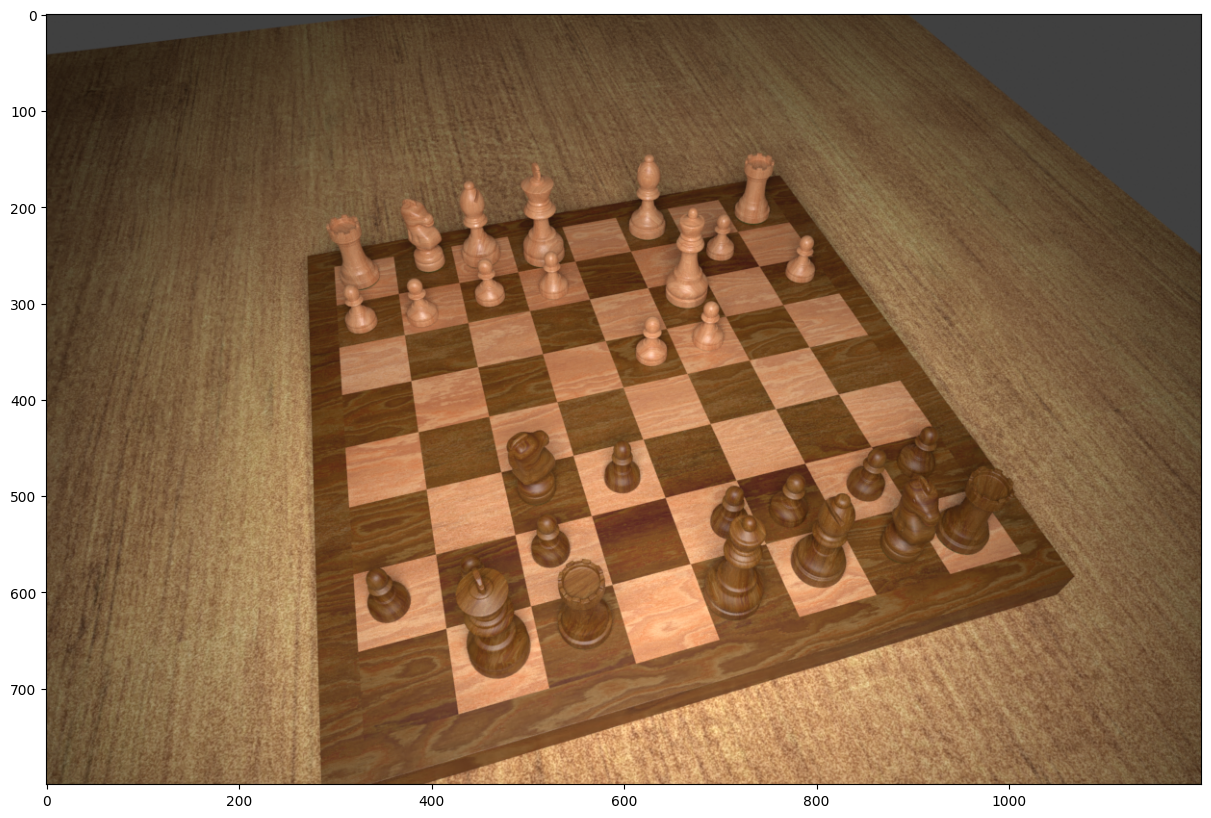}
\caption{Input image.}
\end{subfigure}\hfill
\begin{subfigure}{.36\linewidth}
\includegraphics[width=\linewidth]{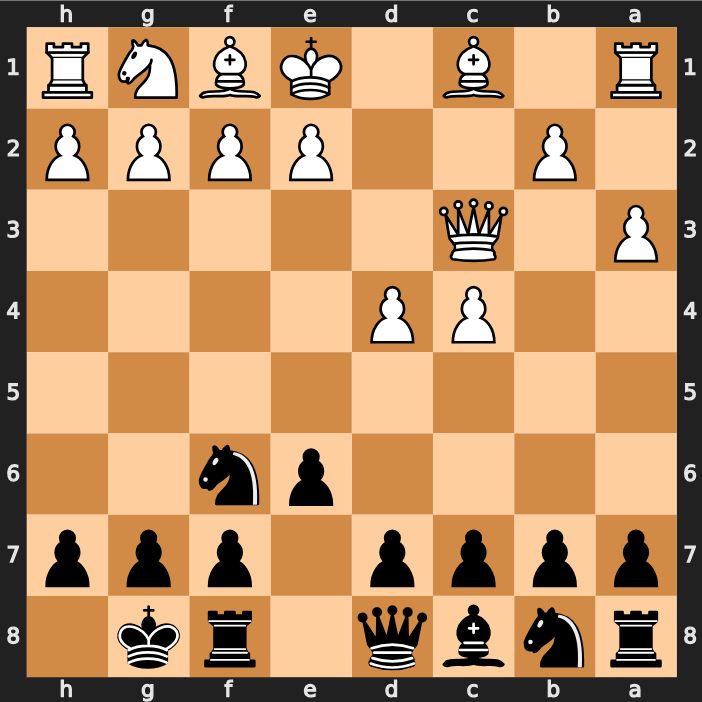}
\caption{Reconstructed position.}
\end{subfigure}

\medskip

\begin{subfigure}{.58\linewidth}
\includegraphics[width=\linewidth]{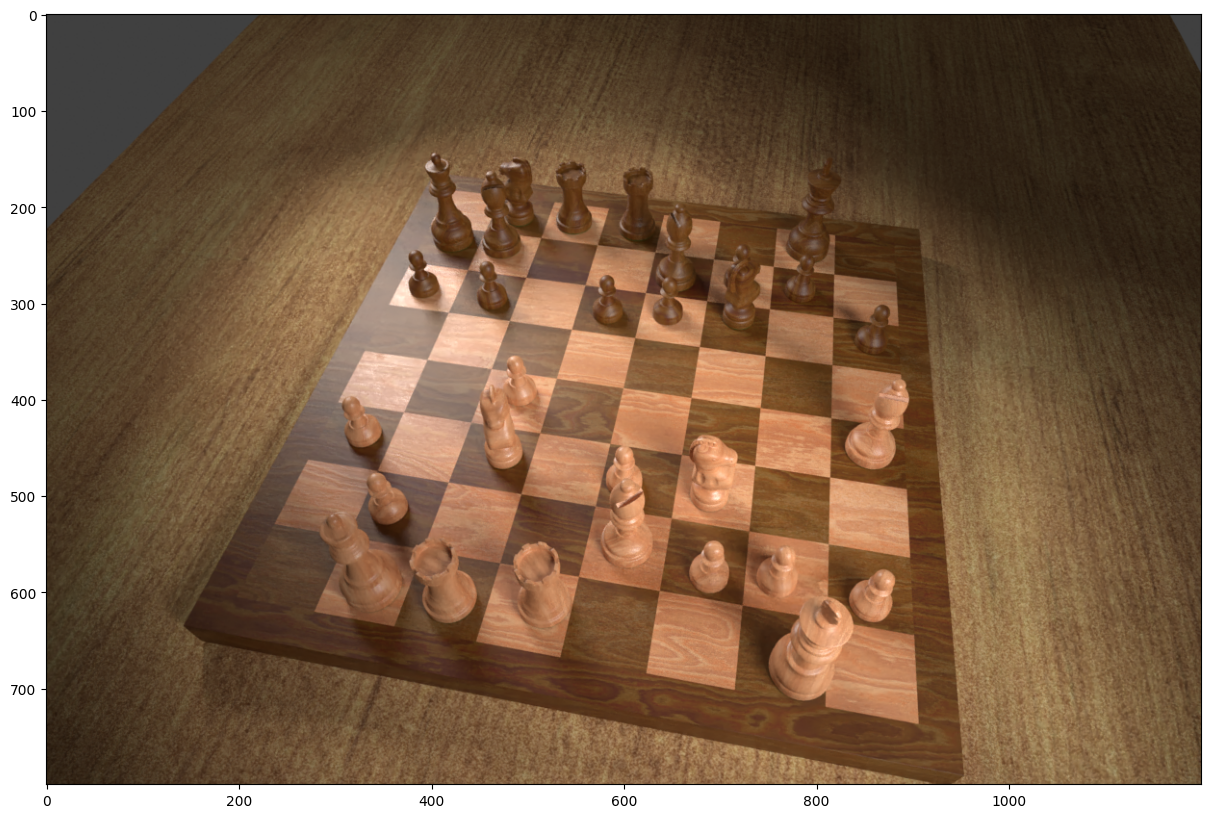}
\caption{Input image.}
\end{subfigure}\hfill
\begin{subfigure}{.36\linewidth}
\includegraphics[width=\linewidth]{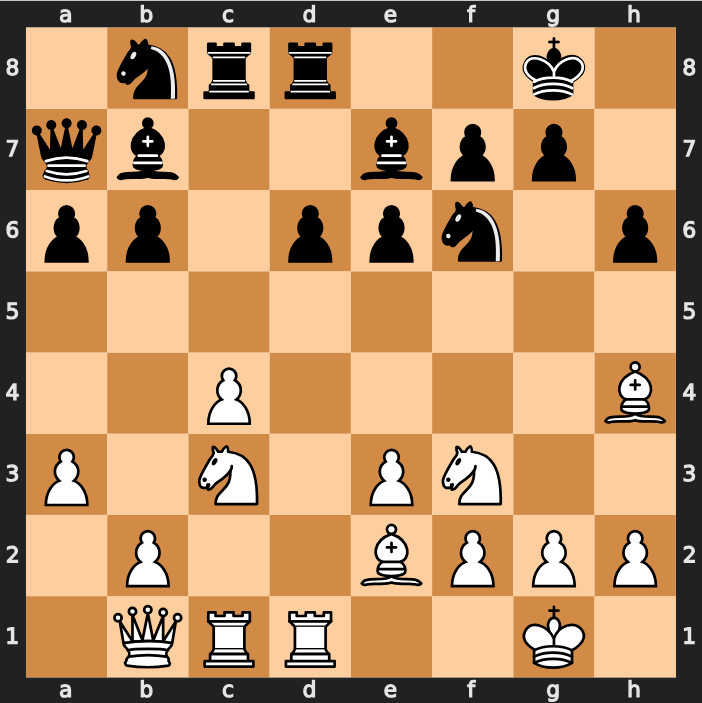}
\caption{Reconstructed position.}
\end{subfigure}
\caption{Examples of complete position reconstruction. The board diagrams
show the system's predictions, with file and rank labels indicating the
display orientation.}
\label{fig:positions}
\end{figure}

The predicted bounding boxes from the piece detection model were projected
using the homography from the board detection model, and the final chess
position was compared square by square with the true position. We report
the mean number of incorrect squares per board, the percentage of boards
with no mistakes, the percentage with at most one mistake, and the per-square
error rate. These measure different aspects of the result. In particular,
one extra pawn makes a complete position incorrect even when the other
63 squares are correct. Figure~\ref{fig:positions} shows examples of input
images and the corresponding reconstructed positions.

Table~\ref{tab:positions} shows the performance of the complete framework.
The test per-square accuracy is 99.57\%, with 76.61\% of boards recovered
without mistakes. Allowing one mistake increases the latter figure to 96.49\%.
The mean number of incorrect squares per board is 0.28. The concentration of
boards with one incorrect square is consistent with the extra pawn detections
observed in the qualitative evaluation.

\begin{table}[tbp]
\centering
\small
\caption{Complete-system results. The final row evaluates square assignment
using true piece boxes and classes with predicted geometry.}
\label{tab:positions}
\begin{tabular}{lrrr}
\toprule
Metric & Training & Validation & Test\\
\midrule
Mean incorrect squares per board & 0.30 & 0.36 & 0.28\\
Boards with no mistakes (\%) & 76.44 & 70.42 & 76.61\\
Boards with at most one mistake (\%) & 95.12 & 94.37 & 96.49\\
Per-square error rate (\%) & 0.47 & 0.56 & 0.43\\
Correct boards with true piece boxes (\%) & 99.74 & 100.00 & 100.00\\
\bottomrule
\end{tabular}
\end{table}

Chesscog was trained on the same source dataset and provides a useful
comparison. Table~\ref{tab:comparison} compares our results with those
published by W\"olflein and Arandjelovi\'c~\cite{Wolflein2021}.
Chesscog achieves 93.86\% complete position accuracy on its test set,
exceeding our accuracy by 17.25 percentage points. With one mistake allowed,
it achieves 99.71\%, compared with our 96.49\%. Thus the aggregate
measurements favour Chesscog for complete reconstruction. The two methods
differ in their input requirements. Chesscog receives the board orientation
as input, whereas our framework predicts labelled correspondences from the
image and uses them to recover both board geometry and orientation.

The methods also organize piece recognition differently. Chesscog first
classifies square occupancy and then classifies the pieces in occupied
squares. In our framework, the piece hypotheses are generated before
assignment to squares, using the class-weighted detector and its deeper
classification head. The retained hypotheses are then resolved through
piece-count limits and geometric assignment. The complete-position results
show the remaining difficulty in converting these detections into an exact
arrangement, particularly when an additional pawn prediction survives the
constraints.

\begin{table}[tbp]
\centering
\small
\caption{Complete position recognition on the synthetic chess dataset.
Chesscog results are from W\"olflein and Arandjelovi\'c~\cite{Wolflein2021}.
Board orientation is provided to Chesscog and predicted by our framework.}
\label{tab:comparison}
\begin{tabular}{lrr}
\toprule
Metric & Chesscog & Present system\\
\midrule
Mean incorrect squares per board & 0.15 & 0.28\\
Boards with no mistakes (\%) & 93.86 & 76.61\\
Boards with at most one mistake (\%) & 99.71 & 96.49\\
Per-square error rate (\%) & 0.23 & 0.43\\
\bottomrule
\end{tabular}
\end{table}

\section{Discussion and conclusions}

The piece-recognition results show the effects of adapting the detector to
the frequencies and visual similarities of chess pieces. Class weighting
raises mAP from 61.59\% to 78.64\%, and the deeper classification head
increases it to 90.14\%. Piece-count filtering improves conditional
classification accuracy by selecting among competing detections, although
it also removes some correctly localized pieces. The proposed procedure
therefore combines changes to the learnt classifier with constraints applied
to its output. Geometric assignment completes this procedure by determining
which predictions compete for the same square.

For piece detection, it has been observed that the model often mistakes the
heads of pieces such as bishops, kings and queens for pawns, sometimes with
high confidence scores. As a result, some final positions have an extra pawn
on the board. A possible reason for this behaviour is related to occlusion.
In some images, a pawn is occluded by a king or queen and, given the camera
angle, the head of the larger piece appears next to the pawn. The model may
have learnt an association between piece heads and the presence of otherwise
occluded pawns. Confusion between similar local shapes provides another
possible explanation. The observed increase in unmatched predictions after
class weighting suggests that further improvements should be evaluated
against false detections as well as classification accuracy.

The choice of keypoints exposes another source of error that is specific to
the detector's representation. Restricting the keypoint head to a predicted
box makes an excluded corner impossible to recover, irrespective of whether
its appearance remains clear in the input image. The octagonal arrangement
provides redundant correspondences and avoids the collinearity of the centre
with opposite outer corners. Every four-point subset is suitable for minimal
homography estimation, giving RANSAC alternative subsets when some predictions
are inaccurate. The labels also carry the board orientation through to square
assignment. These properties explain the choice of representation and its
role in connecting image detections to a chess position.

A useful extension of the framework would be recognition of boards that use
different piece sets from the dataset used for training. Transfer learning
could be used to adapt the detector to pieces from a new dataset. Recognition
of an unseen set without such adaptation remains the more demanding problem,
since the relative appearance of pieces would have to compensate for changes
in their individual features. Variation in piece shape also motivates an
adaptive replacement for the fixed 0.2-square coordinate offset. Estimating
the base of each piece directly would allow square assignment to accommodate
different base sizes and viewing angles.

Board orientation presents a further limitation. The checkerboard pattern
alone leaves two orientations consistent with the requirement that a1 is
dark. Piece distribution can be informative, but is not a reliable rule,
particularly in endgames where pieces may cross to the opponent's half of
the board. The training evaluation included three orientation errors. In the
test images, the labelled keypoint model recovered the orientation required
for correct assignment of all true piece boxes. Board labels, where present,
offer a further source of information. Where the image remains ambiguous,
orientation would have to be confirmed by the user or obtained from
additional information about the game.

We have proposed an R-CNN-based framework for chess position recognition
with complementary contributions to piece recognition and board geometry
estimation. The piece-recognition procedure combines a class-weighted
objective and a deeper detection head with constraints on the predicted
pieces. The board component uses an octagonal arrangement of labelled
keypoints to obtain redundant homography correspondences and recover
orientation. Their predictions are joined through a geometric assignment
procedure that maps piece boxes to board squares. On the synthetic test set,
the final detector achieves a mAP of 90.14\%, and the predicted board
geometry assigns all true piece boxes to the correct squares. The complete
framework recovers 76.61\% of test positions exactly and 96.49\% with at
most one incorrect square. Further development can focus on rejecting pawn
detections associated with the heads of larger pieces and adapting square
assignment to variation in piece bases.


\begin{thebibliography}{15}
\providecommand{\natexlab}[1]{#1}
\providecommand{\url}[1]{\texttt{#1}}
\expandafter\ifx\csname urlstyle\endcsname\relax
  \providecommand{\doi}[1]{doi: #1}\else
  \providecommand{\doi}{doi: \begingroup \urlstyle{rm}\Url}\fi

\bibitem[Czyzewski et~al.(2020)Czyzewski, Laskowski, and Wasik]{Czyzewski2020}
Maciej~A. Czyzewski, Artur Laskowski, and Szymon Wasik.
\newblock Chessboard and chess piece recognition with the support of neural networks.
\newblock \emph{Foundations of Computing and Decision Sciences}, 45\penalty0 (4):\penalty0 257--280, 2020.
\newblock \doi{10.2478/fcds-2020-0014}.

\bibitem[Edwards(1994)]{Edwards1994}
Steven~J. Edwards.
\newblock Portable game notation specification and implementation guide, 1994.
\newblock URL \url{https://www.lokasoft.nl/pgn-standard.aspx}.
\newblock Coordinator; revised 12 March 1994.

\bibitem[{FIDE}(2023)]{FIDE2023}
{FIDE}.
\newblock {FIDE} laws of chess taking effect from 1 january 2023, 2023.
\newblock URL \url{https://handbook.fide.com/chapter/E012023}.

\bibitem[Fischler and Bolles(1981)]{Fischler1981}
Martin~A. Fischler and Robert~C. Bolles.
\newblock Random sample consensus: A paradigm for model fitting with applications to image analysis and automated cartography.
\newblock \emph{Communications of the ACM}, 24\penalty0 (6):\penalty0 381--395, 1981.
\newblock \doi{10.1145/358669.358692}.

\bibitem[He et~al.(2017)He, Gkioxari, Doll{\'a}r, and Girshick]{He2017}
Kaiming He, Georgia Gkioxari, Piotr Doll{\'a}r, and Ross Girshick.
\newblock {Mask R-CNN}.
\newblock In \emph{Proceedings of the IEEE International Conference on Computer Vision (ICCV)}, pages 2961--2969, 2017.
\newblock \doi{10.1109/ICCV.2017.322}.

\bibitem[Lin et~al.(2017)Lin, Doll{\'a}r, Girshick, He, Hariharan, and Belongie]{Lin2017}
Tsung-Yi Lin, Piotr Doll{\'a}r, Ross Girshick, Kaiming He, Bharath Hariharan, and Serge Belongie.
\newblock Feature pyramid networks for object detection.
\newblock In \emph{Proceedings of the IEEE Conference on Computer Vision and Pattern Recognition (CVPR)}, pages 2117--2125, 2017.
\newblock \doi{10.1109/CVPR.2017.106}.

\bibitem[Masouris and van Gemert(2024)]{Masouris2024}
Athanasios Masouris and Jan van Gemert.
\newblock End-to-end chess recognition.
\newblock In \emph{Proceedings of the 19th International Joint Conference on Computer Vision, Imaging and Computer Graphics Theory and Applications}, volume 4: VISAPP, pages 393--403. SciTePress, 2024.
\newblock \doi{10.5220/0012370200003660}.

\bibitem[Ranasinghe et~al.(2023)Ranasinghe, Ranasinghe, and Ashan]{Ranasinghe2023}
Pamudu Ranasinghe, Pasindu Ranasinghe, and Vimukthi Ashan.
\newblock {ChessEye}: An integrated framework for accurate and efficient chessboard reconstruction.
\newblock In \emph{2023 Moratuwa Engineering Research Conference (MERCon)}, pages 177--182. IEEE, 2023.
\newblock \doi{10.1109/MERCon60487.2023.10355515}.

\bibitem[Ren et~al.(2015)Ren, He, Girshick, and Sun]{Ren2015}
Shaoqing Ren, Kaiming He, Ross Girshick, and Jian Sun.
\newblock {Faster R-CNN}: Towards real-time object detection with region proposal networks.
\newblock In \emph{Advances in Neural Information Processing Systems}, volume~28, 2015.
\newblock URL \url{https://papers.neurips.cc/paper_files/paper/2015/hash/14bfa6bb14875e45bba028a21ed38046-Abstract.html}.

\bibitem[Seytre(2026)]{Seytre2026}
Jo{\"e}l Seytre.
\newblock {ChessQueries}: Toward better chess board recognition, 2026.
\newblock URL \url{https://arxiv.org/abs/2608.30762}.
\newblock arXiv preprint.

\bibitem[W{\"o}lflein and Arandjelovi{\'c}(2021)]{Wolflein2021}
Georg W{\"o}lflein and Ognjen Arandjelovi{\'c}.
\newblock Determining chess game state from an image.
\newblock \emph{Journal of Imaging}, 7\penalty0 (6):\penalty0 94, 2021.
\newblock \doi{10.3390/jimaging7060094}.

\bibitem[Wu et~al.(2019)Wu, Kirillov, Massa, Lo, and Girshick]{Wu2019}
Yuxin Wu, Alexander Kirillov, Francisco Massa, Wan-Yen Lo, and Ross Girshick.
\newblock {Detectron2}.
\newblock \url{https://github.com/facebookresearch/detectron2}, 2019.

\bibitem[Xie et~al.(2017)Xie, Girshick, Doll{\'a}r, Tu, and He]{Xie2017}
Saining Xie, Ross Girshick, Piotr Doll{\'a}r, Zhuowen Tu, and Kaiming He.
\newblock Aggregated residual transformations for deep neural networks.
\newblock In \emph{Proceedings of the IEEE Conference on Computer Vision and Pattern Recognition (CVPR)}, pages 1492--1500, 2017.
\newblock \doi{10.1109/CVPR.2017.634}.

\bibitem[Xie et~al.(2018)Xie, Tang, and Hoff]{Xie2018Chess}
Youye Xie, Gongguo Tang, and William Hoff.
\newblock Chess piece recognition using oriented chamfer matching with a comparison to {CNN}.
\newblock In \emph{2018 IEEE Winter Conference on Applications of Computer Vision (WACV)}, pages 2001--2009, 2018.
\newblock \doi{10.1109/WACV.2018.00221}.

\bibitem[Zhu et~al.(2023)Zhu, Wong, and McPhee]{Zhu2023}
Kevin Zhu, Alexander Wong, and John McPhee.
\newblock {VICE}: View-invariant chess estimation.
\newblock In \emph{Proceedings of the 11th International Conference on Sport Sciences Research and Technology Support (icSPORTS)}, pages 50--60. SciTePress, 2023.
\newblock \doi{10.5220/0012167200003587}.

\end{thebibliography}
\end{document}